\documentclass{article}
\usepackage{iclr2027_conference,times}
\usepackage[T1]{fontenc}
\usepackage[utf8]{inputenc}
\usepackage{graphicx}
\usepackage{booktabs}
\usepackage{multirow}
\usepackage{array}
\usepackage{amsmath}
\usepackage{amssymb}
\usepackage{subcaption}
\usepackage{tikz}
\usetikzlibrary{arrows.meta,positioning,shapes.misc}
\usepackage[hyphens]{url}
\usepackage[hidelinks]{hyperref}
\usepackage{microtype}
\usepackage{xcolor}

\newcommand{\gw}{SemDraft}
\newcommand{\fullb}{the development corpus}
\newcommand{\tight}{\emph{tight}}
\newcommand{\cert}{\emph{certified}}

\title{Learning to Search for Executable\\Patternmaking from Certified Programs}

\author{%
  \textbf{Zhuotao Xu, Chi Zeng, Liying Yang, Xinxing Yu, Yanyan Liang} \\
  Macau University of Science and Technology%
}

\iclrfinalcopy

\begin{document}
\maketitle
\lhead{Preprint}

\begin{abstract}
Recent models generate sewing patterns from images or text, and program-based representations make these patterns executable. These systems produce a finished pattern and do not model how a patternmaker turns one pattern into another. In practice a new style is made by editing an existing pattern: lengthening the bodice, moving a dart, swapping a sleeve. Pattern archives record the results of such edits and none of the steps. We observe that the steps can be recovered by searching over editing actions until one pattern becomes the other. A deterministic compiler makes the recovered steps checkable. To this end, we present \gw{}, the first framework that learns to search for executable patternmaking programs from paired before-and-after sewing patterns without recorded editing steps. Specifically, a bounded search recovers a program of editing actions and certifies it only after three checks: it recompiles from scratch, every pair of its actions is classified as ordered or interchangeable, and an independent replay reaches the same pattern. The certified programs then train a model that ranks candidate actions inside the same search. On two frozen draws of 80 order-sensitive cases, with every other search component fixed, the learned ranker certifies 63 and 64 cases within 1,000 search executions, against at most 53 and 52 for non-learned rankers.
\end{abstract}

\section{Introduction}
\label{sec:intro}

Sewing patterns underlie digital garment creation, from virtual try-on to made-to-measure manufacturing.
Recent systems generate complete patterns from images or text \citep{liu2023sewformer,he2024dresscode,bian2025chatgarment,nakayama2025aipparel,weng2026garmentgpt}, and program-based representations make these patterns executable \citep{korosteleva2023garmentcode,zhou2025design2garmentcode}.
Patternmakers rarely draft a style from nothing.
They edit an existing pattern by lengthening a bodice, adding and transferring darts, redistributing ease, or swapping a sleeve template.
The expertise that makes garments fit is therefore stored only as its outcomes.
We study this setting: given the pattern before the edits and the pattern after, recover an executable program of editing actions that connects them.

Recovering the editing actions is challenging for three reasons.
First, the supervision is missing.
Even when a base--target pair is available, industrial CAD archives keep no log of the operations that connect them.
Second, a garment action is coupled.
One action changes several panels, their seam partners and their symmetry groups at once, whereas the operations recovered by inverse CAD methods act on a single solid \citep{xu2021zonegraphs,khan2024cadsignet,dupont2024transcad,rukhovich2025cadrecode}.
Third, order is ambiguous.
Many orders of the same actions produce the same pattern, and some orders do not, because an attached sleeve template resolves its geometry against the host panels as they are at that moment.
A single recovered sequence therefore leaves open which other orders of its actions are valid.

\begin{figure}[t]
\centering
\includegraphics[width=0.94\linewidth]{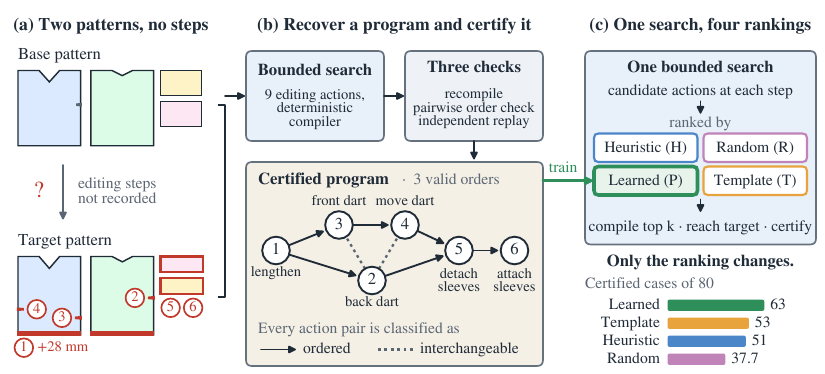}
\caption{Overview of \gw{} on one case of the development corpus. (a) The input is a base pattern and a target pattern; the editing steps between them are not recorded. Red marks the changes, numbered by the step that makes them. (b) A bounded search over nine editing actions, executed by a deterministic compiler, recovers a program, and three checks certify it; the certificate classifies every pair of actions as ordered or interchangeable, which here admits three orders. (c) Certified programs train a model that ranks candidate actions (P). It is compared with the teacher heuristic (H), a random order (R) and a template lookup (T) inside one bounded search with the same candidates, beam, stopping rule, certifier and budget; bars: certified main-test cases of 80 at beam 4.}
\label{fig:overview}
\end{figure}

Our insight is that a deterministic compiler makes a recovered sequence checkable.
Once a compiler executes the editing actions, a program that reaches the target can be recompiled from scratch. Each pair of its actions can then be classified as ordered or interchangeable, and an independent replay can confirm the result.
We build \gw{} on this insight (Fig.~\ref{fig:overview}).
A drafting environment compiles nine semantic patternmaking actions to geometry.
A bounded teacher search recovers programs and accepts only those that pass the three-stage certification.
The certified programs train a learned action-selection prior that ranks the environment's legal candidates.
Every training program is therefore executable and carries a certified dependency DAG, with the valid next actions at each intermediate state.
A unified bounded search then runs every ranking signal, learned or not, with the same candidates, beam, stopping rule, certifier and a shared execution budget.

We evaluate on a newly generated and frozen test set of 80 order-sensitive cases, each requiring a template swap whose landing depends on the host panels.
Under matched conditions, the learned prior with a beam of four certifies 63 of the 80 cases, averaged over two training seeds.
A lookup of the training corpus's action-type statistics certifies 53, the teacher's own ranking heuristic 51, and a random order 37.7.
The comparisons are paired on the same cases, the bootstrap range of each gain lies above zero, and an independent draw of 80 further cases reproduces them.
The teacher itself, run unchanged on the same cases with its full wall-clock budget, certifies 30.
Of the 50 cases it misses, the prior certifies 34 and the teacher heuristic 21, each within 1,000 search executions.

In summary, our main contributions are:
\begin{itemize}
\item We present the first framework that learns to search for executable patternmaking programs from paired before-and-after sewing patterns without recorded editing steps. A bounded search over semantic patternmaking actions accepts a program only after a three-stage certification: recompilation, a pairwise order check and an independent replay. Every training program is therefore executable and carries its certified dependency DAG and the valid next actions at each of its intermediate states.
\item We propose a learned action-selection prior and evaluate it under matched conditions. The prior ranks the environment's legal candidates inside one fixed bounded search. We compare it with the teacher's heuristic, a random order and a template lookup under the same candidates, beams, stopping rule, certifier and execution budgets.
\item We establish a frozen evaluation with replication, distribution shift and teacher-missed solutions. The test set is freshly generated, de-duplicated and split before use, with an independent replication draw and two distribution-shift sets. On the cases the teacher itself misses within its wall-clock budget, we test whether the prior finds certified programs.
\end{itemize}

\section{Related Work}
\label{sec:related}

\paragraph{Sewing-pattern generation and alteration.}
This line produces or modifies complete sewing patterns.
GarmentCode and GarmentCodeData define parametric, executable pattern programs and a made-to-measure dataset \citep{korosteleva2023garmentcode,korosteleva2024garmentcodedata}.
SewFormer, DressCode, ChatGarment, AIpparel, Design2GarmentCode and GarmentGPT generate or edit patterns from images, text or both \citep{liu2023sewformer,he2024dresscode,bian2025chatgarment,nakayama2025aipparel,zhou2025design2garmentcode,weng2026garmentgpt}, and GarmentGPT replaces coordinate regression with discrete, compositional latent tokens.
Digital garment alteration is the closest formulation to ours: it edits an existing pattern with tailoring operations and fits it to a new body shape \citep{eggler2024garmentalteration}.
Concurrently, TailorCoPilot records the operations that experts perform in a pattern-making tool as explicit sequences over structured pattern states \citep{sun2026tailorcopilot}.
\gw{} recovers such sequences from the two patterns alone and learns to choose the next action, anchor and amount.

\paragraph{Recovering construction sequences and programs.}
Inverse CAD recovers how an object was built.
With recorded histories, DeepCAD, SkexGen and Text2CAD model ordered command sequences \citep{wu2021deepcad,xu2022skexgen,khan2024text2cad}.
From final geometry, Zone Graphs, PLAD, CAD-SIGNet, TransCAD and CAD-Recode infer sequences or code \citep{xu2021zonegraphs,jones2022plad,khan2024cadsignet,dupont2024transcad,rukhovich2025cadrecode}, ExtrudeNet, SECAD-Net and SfmCAD learn sketch-and-extrude decompositions without them \citep{ren2022extrudenet,li2023secadnet,li2024sfmcad}, and cadrille adds reinforcement learning from programmatic feedback \citep{kolodiazhnyi2026cadrille}.
VPI-Edit learns local edits that move a visual program towards a target and evolves a population of programs with them \citep{jones2024vpiedit}.
CAD-Assistant applies tool-augmented VLMs to CAD tasks \citep{mallis2025cadassistant}, and execution-guided synthesis trains a learner on programs found with an executor in the loop \citep{ellis2019write,ellis2021dreamcoder}.
These methods reconstruct a single shape from its geometry.
A garment action changes several panels and their seams, and many orders of the same actions reach the same pattern.
\gw{} recovers a certified partial order of parameterised actions over coupled panels, with one executor for training and evaluation.

\paragraph{Learning from search and verification.}
A common pattern trains a model on the outputs of a stronger search and then uses it to guide a cheaper search.
Expert iteration, AlphaGo Zero, pseudo-label program induction such as PLAD, and library learning in DreamCoder follow this pattern \citep{anthony2017thinking,silver2017alphago,jones2022plad,ellis2021dreamcoder}.
In program synthesis, learned models order a fixed enumeration or deduction \citep{balog2017deepcoder,kalyan2018neural} or act as the search policy \citep{shi2022crossbeam}, and a learned policy can improve on a suboptimal reference \citep{chang2015learning}.
Verifier-filtered self-training is common in reasoning tasks \citep{cobbe2021verifiers,singh2024beyond}.
When several next actions are equally correct, dynamic oracles supply the set of optimal transitions \citep{goldberg2013training}, and training can maximise their total probability \citep{ballesteros2016training}, the objective we use to train the prior.
Deordering a sequential plan into a partial order is classical in planning \citep{muise2016optimal}, and the equivalence of interleavings under an independence relation is the subject of trace theory \citep{mazurkiewicz1987trace}. A concurrent system certifies programs recovered from execution video with a checker over such an independence quotient \citep{si2026checked}.
\gw{} decides independence by executing both orders in a geometric compiler and keeps a program only after all three checks succeed. It trains the prior to place its probability mass on the valid next actions that the certificates provide.

\section{Method}
\label{sec:method}

Given a base pattern $S_0$ and a public target, \gw{} searches for an executable program that connects them. A compiler executes the actions, a teacher search recovers and certifies programs, and a learned prior ranks candidates inside the same search.

\subsection{Drafting environment: semantic actions over a deterministic compiler}
\label{sec:env}

\paragraph{State and actions.}
A pattern state $S$ has panels (closed polygons with semantically labelled edges), seams, components such as a bodice or sleeve pair, and internal darts as separate entities.
Edits use the vocabulary of patternmaking. The ontology has nine semantic actions: introduce internal dart, transfer internal dart, redistribute component ease, true a straight seam pair, adjust component length, apply axis slash-and-spread, insert style line, and attach or detach a finite template component.
Each action names the panels, boundaries, darts or components it touches, and carries continuous parameters, given in millimetres or in normalised units.

\paragraph{Compiler.}
One semantic action changes several panels, their seam partners and their symmetry groups together, so actions are executed by a compiler.
The compiler lowers every action to a fixed set of geometric primitives (lengthen/shorten, widen/narrow, reshape boundary, slash-spread, split, merge, attach, detach).
Execution is deterministic and transactional, with stable entity identities and full lineage.
A transition that breaks closure, self-intersects, or violates seam compatibility or dart validity is rejected.
Determinism makes a recovered program checkable, because the same program always reproduces the same state and artefact hash.
Execution is also order-sensitive: attaching a template component such as a sleeve resolves its geometry against the current host panels, so the same swap yields different sleeves before and after a length adjustment (Fig.~\ref{fig:walkthrough}).
The compiler is fixed throughout the paper, and learning enters only through the ranking of candidates.

\paragraph{Public endpoint criterion.}
The teacher and the prior observe a public target: panel contours, boundary semantics, dart parameters and component tags.
The generating program stays hidden.
A state is \tight{} with respect to the target when three conditions hold.
Eight exact structure predicates agree with the target: components, panels, boundary semantics, topology, seam graph, darts, style-line tags and template tags.
A complete correspondence exists between the entities of the state and those of the target.
Every panel contour lies within fixed public thresholds of its matched target contour: Chamfer $\le 0.005$, Hausdorff $\le 0.020$ and aligned RMS $\le 0.010$. Each distance is normalised by the centroid and bounding-box diagonal of the target contour and measured on 512 samples per contour.

\subsection{Teacher: bounded search and certification}
\label{sec:teacher}

\paragraph{Search.}
Archives provide no programs, so \gw{} first recovers them with a teacher, a bounded search whose acceptance test is a three-stage certification.
Given $S_0$ and the public target, the teacher runs a beam search over programs of bounded depth.
At each prefix, nine proposal providers, one per action type, enumerate candidate actions that are consistent with the state--target correspondence.
A dart present in the target and absent in the state yields a dart proposal.
Every candidate is compiled and evaluated with the public metric.
Prefixes are ranked lexicographically by structural mismatches, then contour distance, then program cost, and a beam of $B$ prefixes survives per depth.
The search is bounded in wall clock and in engine executions (Appendix~\ref{app:protocol}).
Two mechanisms serve template swaps, which every case of the hardest benchmark tier requires: a one-step lookahead from each proposed detach to its re-attach variants, and a certification reserve of wall-clock time.
Both mechanisms were developed using an aggregate census of the hidden programs of \fullb{} and form part of the teacher's reference configuration.
This configuration is run separately and unchanged; the unified search uses its ranking rule within a shared search loop.

\paragraph{Certification.}
A prefix whose endpoint is tight is a terminal candidate.
It becomes \cert{} when three independent checks succeed.
The first check recompiles the whole program from $S_0$, which must reproduce the terminal state and its artefact hash.
The second, a pairwise order check, gives every pair of actions one verdict.
A pair whose static read/write effects conflict is recorded as a required precedence without execution.
A pair that becomes adjacent in some explored order is executed in both orders from the shared prefix, and the artefact hashes are compared.
Equal hashes give a commuting pair, two valid and different results give a required precedence, and a side that cannot be materialised leaves the pair undecided.
Each pair is decided once and cached, and a pair that never becomes adjacent keeps the recovered order as a precedence implied by those already established.
The check is complete when the exploration of commuting swaps closes with no undecided pair.
The certifier then enumerates the reachable linear orders, materialises every downward-closed subset of the resulting dependency DAG as an executable intermediate state, and applies a minimality rule: no proper downward-closed subset may already be tight.
The third check, an independent replay, runs the compiled low-level program in an oracle-isolated evaluator, verifies lineage completeness and locality, and confirms that the endpoint is tight under the hidden full state.
Appendix~\ref{app:cert} states what a certificate establishes and where it is conservative.
A certificate is the unit of supervision: it yields the program, its dependency DAG and, for every materialised intermediate state, the set of \emph{valid next actions} (nodes whose required predecessors are complete) and whether stopping is allowed there.

\subsection{Learned prior: a factorised model trained on certified programs}
\label{sec:student}

\paragraph{Model.}
The teacher ranks a candidate only after compiling it; the prior ranks candidates before compilation.
It observes what the teacher observes: the panels, darts and components of the current state, the panels and darts of the public target, the state--target correspondence, the public per-panel metrics, and the actions executed so far.
Each panel, target panel, component, dart and past action becomes a token with geometric statistics, and a global token carries the aggregate metrics and mismatch counts (Appendix~\ref{app:policy}).
A four-layer Transformer encoder \citep{vaswani2017attention} with hidden dimension 128 and four attention heads feeds three output heads. A type head scores the nine action types and STOP, an anchor head points to a panel or component token given the type, and a parameter head regresses millimetre amounts and classifies discrete slots.

\paragraph{Training.}
For every materialised state $s$ of a certified program, the positive set $\mathcal{P}(s)$ holds all valid next actions, together with STOP where stopping is allowed.
With $p(a\mid s)$ the prior's probability of action $a$ in $s$, the loss is
\begin{equation}
\mathcal{L}(s) = -\log \sum_{a\in\mathcal{P}(s)} p(a\mid s),
\label{eq:dag-loss}
\end{equation}
applied to the type and to the anchor given each positive type, plus regression and classification losses on the parameters of the positives.
We call this \emph{DAG supervision}.
The ablation \emph{linear imitation} uses the same model and data and restricts $\mathcal{P}(s)$ in Eq.~\eqref{eq:dag-loss} to the next action of the terminal's linear order. This is how behaviour cloning would treat a recovered sequence.

\subsection{Unified bounded search: one search, four ranking signals}
\label{sec:unified}

\paragraph{Search loop.}
To measure what a ranking signal contributes, every other component of the search is held fixed.
At each prefix the nine providers enumerate the legal candidates as for the teacher, up to eight per provider.
A ranker orders the candidates, and they are compiled in that order until $k$ of them compile ($k\in\{1,4\}$).
The $k$ best children over all expanded prefixes form the next beam, where a child is ranked by the signal that ordered it.
The search ends when a prefix is tight, when the depth limit of eight is reached, or when the execution budget is exhausted.
All ranking signals share the public tight stopping rule, isolating the contribution of candidate ranking.
The budget counts the engine executions spent in the search, which we call search executions. Each compilation is charged the executions it uses, and the budget is set as a fraction of the teacher's per-tier execution cap.
Certification is charged separately, with the tier's cap as its limit.

\paragraph{Ranking signals.}
The heuristic H is the teacher's own ranking rule.
It compiles every candidate and ranks the compiled states by the public metric key, as the teacher does inside its beam.
The random order R is averaged over three seeds.
The template lookup T orders candidates by the empirical probability of their action type given the tier and the step index in the certified training programs, and uses no geometry.
The learned prior P is trained with DAG supervision or linear imitation.
It scores a candidate by the log-probabilities of its action type and of its anchor given that type, plus a term that rewards amounts close to those the prior predicts (Eq.~\eqref{eq:score} in Appendix~\ref{app:policy}). A child inherits the accumulated score of its prefix.
H pays for its ranking information by compiling all candidates at every step, whereas R, T and P compile only the candidates that they select.
A run achieves \emph{certified success} at budget $b$ if its first tight program is found within $b$ search executions and passes the three-stage certification.
The choices of a run are independent of the budget, so one run per condition yields the whole success-versus-budget curve.

\section{Benchmarks: a development corpus and frozen evaluation sets}
\label{sec:benchmark}

\paragraph{Generator.}
All corpora come from one generator.
Each case starts from the neutral base pattern of its family (tee, skirt, pants or dress, each with a base dart and a detachable template component) and applies a hidden program to obtain the target.
Programs are templated: the generator fixes the action-type order for each tier and depth, and a hash of the corpus label and the case slot picks the anchors and amounts from small value sets.
Tier A programs have depth 2--3, Tier B 4--5 and Tier C 6--8.
Every Tier-C program contains a detach/attach template swap, so the position of the swap relative to the length, dart and ease actions is part of the problem.
Templating has two consequences.
The action-type sequence of a case is largely determined by its tier and depth, and the template lookup T tests how far this knowledge alone goes.
What remains for the search is the anchor and amount of each action and, at Tier C, an order in which the swap follows the host changes it depends on.
At run time the teacher, the prior and the evaluation scripts receive public fields only, and a code-level guard rejects any oracle field that reaches the search interface.

\paragraph{Development corpus.}
Our development corpus contains 300 cases, 25 per tier and family; 5 cases per tier and family were used to develop the teacher and 20 are held out.
It is the benchmark on which the teacher was developed, and it serves three purposes here: its certified programs train the prior, it supports protocol development, and it provides diagnostics.
Teacher budgets are fixed per tier.

\paragraph{Frozen evaluation sets.}
For the main comparison we generated a new corpus of 300 cases with the same generator under a new corpus label.
It was frozen before any ranker ran on it, as recorded in Appendix~\ref{app:protocol}.
Five of its 300 targets overlap with \fullb{}; none belongs to the declared splits below.
Three splits were declared before use.
24 calibration cases, two per tier and family, fix the budgets and beam widths.
The main test is the 80 held-out Tier-C cases (20 per family), and it carries the primary comparison.
An easy reference of 20 held-out Tier-B cases shows the regime where the candidate generators suffice.
Three further sets test robustness: an independent replication draw of 80 Tier-C cases, and a parameter shift and a template shift of 20 cases each. They were added after the main results had been read and fixed before use.
The teacher's reference configuration, unchanged from the development-corpus runs, is also run on the whole main test, and the cases it misses within its wall-clock budget define the evaluation of Sec.~\ref{sec:beyond}.
Every experiment follows a written protocol whose budgets, seeds and gates were fixed before its results were read \citep{nosek2018preregistration}, and Appendix~\ref{app:protocol} lists the hardware and software and logs all deviations.

\section{Experiments}
\label{sec:experiments}

\begin{figure}[t]
\centering
\includegraphics[width=0.84\linewidth]{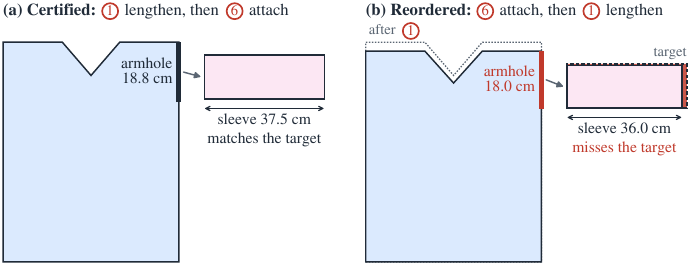}
\caption{In the case of Fig.~\ref{fig:overview}, the sleeve attachment has to follow the length adjustment, because the compiler sets a sleeve's width to the combined length of the front and back armholes it is sewn to. Compiled front bodice and right sleeve at one scale, without the darts, which end up identical in both orders; solid outlines show the state when the sleeve is attached. (a) In the certified order the sleeve matches the target. (b) Attached first, the sleeve is sized from the 18.0\,cm armhole and keeps that size after the bodice is lengthened (dotted), so the program misses the target (Appendix~\ref{app:cert}).}
\label{fig:walkthrough}
\end{figure}

\subsection{Supervision: certified programs and the trained prior}
\label{sec:teacher-results}

\paragraph{Teacher on the development corpus.}
The teacher certifies 77, 70 and 57 of the 80 cases of Tiers A, B and C (Table~\ref{tab:teacher}, Fig.~\ref{fig:teacher}).
On Tier C every tight case is also certified, and certified cases use a median of 1,442 engine executions and 1,315 s of wall clock.
Most of the 23 non-certified Tier-C cases show \emph{beam collapse}. The beam fills with reorderings of the same one or two sets of actions, all reaching the same state, and loses the branch that leads to the target (Table~\ref{tab:failures}).

\paragraph{Training data and teacher-forced accuracy.}
\label{sec:student-results}
The training data are the teacher's certified programs on \fullb{}, 1,510 certified terminals over 185 cases: 77 in Tier A, 51 in Tier B and 57 in Tier C.
They yield 2,846 materialised states after de-duplication by (case, completed action set), each with its set of valid next actions.
A validation split of 20\% of the cases per tier contributes no training states. Teacher-forced on this split, the prior predicts a valid next action type 92\% of the time against a 49\% majority baseline, and the anchor entity 99\% of the time.
Its millimetre amounts halve the error of a mean predictor.
Linear imitation is indistinguishable on these metrics, as expected when the extra positives rarely change the predicted argmax (Table~\ref{tab:tf}).

\subsection{Matched-condition comparison on the frozen evaluation set}
\label{sec:main-results}

The primary endpoint, fixed before the runs, is certified success within 1,000 search executions (5\% of the Tier-C cap) with a beam of four, and beam 1 is secondary.
Two rankers are compared on the same cases, and Table~\ref{tab:paired} gives the case-bootstrap range of every gain.

\begin{figure}[t]
\centering
\includegraphics[width=0.88\linewidth]{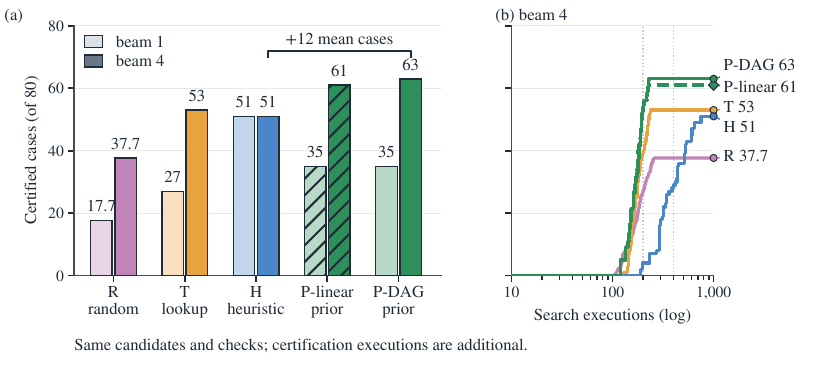}
\caption{Matched-condition comparison on the 80 Tier-C main-test cases. (a) Certified cases per ranking signal at beam 1 (light) and beam 4 (dark) within 1,000 search executions, the 5\% budget; hatching marks the linear-imitation variant of the prior. (b) Certified cases against search executions at beam 4, read off the same runs; dotted lines mark 200 and 400 executions (1\% and 2\% of the cap).}
\label{fig:policy}
\end{figure}

\paragraph{Main result.}
With a beam of four, the learned prior certifies more cases than any non-learned ranking signal (Fig.~\ref{fig:policy} and Table~\ref{tab:main}).
P certifies 63 of the 80 main-test cases, the mean of its two training seeds, 61 and 65.
The template lookup T certifies 53, the teacher heuristic H 51 and the random order R 37.7, and P with linear imitation certifies 61.
The gain over H is 12 cases, or 15 percentage points, and meets the criterion fixed before the runs: at least ten points, with a bootstrap range above zero.
The gains over T and R, 10 and 25 cases, also have ranges above zero.
T knows the action-type statistics of every tier and step and certifies 27 cases at beam 1 and 53 at beam 4, most of them at depths 6 and 7.
Twelve cases are certified by P with both seeds and not by T, and two by T alone. At depth 8 T certifies 5 of 24 cases, against 10 for P and for H (Table~\ref{tab:cells}).

\paragraph{Beam width and search cost.}
H reaches its 51 cases with a median of 103 search executions at beam 1 and 351 at beam 4, and the wider beam adds no case.
At beam 1 P commits to one candidate per step and certifies 35 cases, fewer than H, because the metric feedback of H corrects a wrong first choice.
On the 44 cases that both certify at beam 4, P needs a median of 160 search executions and H 325, and the median per-case ratio is 1.9. On the 49 such cases of the replication draw it is 1.8.
A run ends when its beam is exhausted, after about 220 executions for P.
P, R and T therefore gain no cases after 400 executions. H at beam 4 keeps rising, from 29 cases at 400 executions to 51 at 1,000, and no ranker changes between 1,000 and 4,000 (Fig.~\ref{fig:policy}b).

\paragraph{Tight and certified.}
Eight of P's tight endpoints at beam 4, nine of T's at beam 1 and one of H's fail certification.
All are eight-step programs that reach the target with one redundant step, and the certifier's minimality rule rejects them (Appendix~\ref{app:cert}).
On the easy reference every signal is tight on all 20 cases at beam 1, and 16 to 20 certify, so the differences belong to the order-sensitive tier.

\subsection{Where the gain comes from}
\label{sec:where}

Against H, the gain comes from tee and dress cases at depth 7.
Table~\ref{tab:cells} splits the main test by family and target depth.
At depth 6 both signals certify almost every case: P all 28 and H 25.
At depth 7 P certifies all seven tee cases and all seven dress cases, and H one of each; both certify the seven skirt cases.
At depth 8 the two are level at 10 of 24, and this depth remains open.
Pants is the family on which H is stronger, with 18 of 20 cases against 11.
P misses six pants cases at depth 8, where its beam is exhausted before a tight state, and three at depth 7 that reach the target with one redundant step.
Of H's 29 misses, 27 are tee and dress cases, the families in which the teacher's beam fills with reorderings of the same actions on the development corpus.

\paragraph{Supervision form and beam width.}
DAG supervision and linear imitation are level: 35 against 35 at beam 1, and 63 against 61 at beam 4, within the spread of the two seeds, 61 and 65.
On this templated corpus the sets of valid next actions add little to the single linear order, and the certificate serves as the acceptance test and as the filter for training programs.
The beam width matters more: it raises P from 35 to 63, T from 27 to 53 and R from 17.7 to 37.7, and leaves H unchanged at 51.

\subsection{Robustness: a second draw, shifted inputs and beam de-duplication}
\label{sec:robust}

\begin{figure}[t]
\centering
\includegraphics[width=0.76\linewidth]{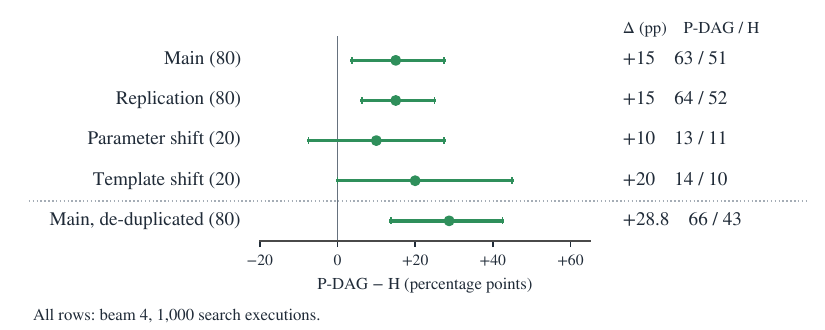}
\caption{Paired gain of the prior (P-DAG) over the teacher heuristic (H) at beam 4 within 1,000 search executions, in percentage points; bars show the case-bootstrap range of each gain. The last row applies beam de-duplication to both. The right column gives the gain and both counts.}
\label{fig:robustness}
\end{figure}

\paragraph{Replication.}
The result reproduces on a second draw of 80 cases, run with the same trained prior and protocol (Fig.~\ref{fig:robustness} and Table~\ref{tab:shift}).
P certifies 64 cases at beam 4 with both seeds and with linear imitation, against 52 for H, 51 for T and 37.3 for R.
The gain over H is again 12 cases, and the gain over T is 13 cases.
Both gains have bootstrap ranges above zero, as does the gain over H of 24 cases on the 160 cases of both draws.
At beam 1 the main-test order repeats: H certifies 52, P 46 and T 33.

\paragraph{Distribution shift within the ontology.}
The learned prior certifies the most cases under both controlled shifts.
The \emph{parameter-range shift} keeps the templates and moves every parameter value set to unseen values.
The \emph{template shift} keeps the parameter values and changes the action composition and dependency structure at all three depths.
Both are evaluated without tuning.
At beam 4 P certifies 13 cases under the parameter shift and 14 under the template shift, against 11 and 10 for H, 9 and 10 for T, and 9.7 and 9.3 for R.
With twenty cases per set, the bootstrap ranges of the gains over H, 2 and 4 cases, include zero, and those of the gains over T, 4 cases on each set, lie above zero.
Under shift the certified rate moves from 79\% to 65--70\% for P, from 64\% to 50--55\% for H and from 66\% to 45--50\% for T. That P loses the least argues against template memorisation.

\paragraph{Beam de-duplication.}
Removing the heuristic's duplicate beam slots widens the gap.
Prefixes that reach the same full state through the same actions in a different order cause the teacher's beam collapse, and they can be merged to free beam slots.
We added this merging to the unified search for H and for P at beam 4 after the main results had been read, and fixed the check before its run.
P gains three cases, from 63 to 66 with seeds 65 and 67, and linear imitation rises from 61 to 64. H loses eight cases, from 51 to 43.
H's beam held a median of 12 merged states per case against 2 for P, so the merging mostly changes H's search.
With merging on both sides P certifies 23 more cases than H, so the learned gain is not explained by the beam slots that H wastes on reorderings.

\subsection{Beyond the teacher's reference configuration}
\label{sec:beyond}

The prior certifies programs that the teacher's reference configuration does not find within its wall-clock budget.
It was run on all 80 main-test cases with its configuration and analysis rule fixed in advance.
The configuration certifies 30 cases (Table~\ref{tab:c3}).
Of the other 50, 5 reach a tight endpoint that fails certification, 27 exhaust the search envelope, and 18 are stopped while writing their result after the search had ended.
All 27 envelope cases stop at the wall-clock limit, far below the execution cap (median 1,482 of 20,000 executions).
On \fullb{} it certifies 57 of 80; the drop to 30 has the same main mechanism, beam collapse.
On the 50 missed cases, P at beam 4 certifies 34 with both seeds within 1,000 executions, T 29, H 21 and R 14.
On the 32 cases not stopped while writing their result, the counts are 25, 20 and 15 for P, T and H.
Seventeen cases are certified by P and not by H, and four the other way; all five programs that the reference configuration reached without certifying are certified by both.
The advantage is at depths 6 and 7, where P certifies 11 of 11 and 18 of 18 cases against 8 and 6 for H. At depth 8 both are weak, with 5 and 7 of 21.

\section{Limitations}
\label{sec:limitations}

The development corpus and the frozen evaluation set are controlled, synthetic identifiability benchmarks, not evidence of industrial generalisation. Their hidden programs are drawn from the same nine-action ontology the search uses, the programs are templated per tier and depth, and the geometry is polygonal.
The teacher's reference configuration has two identified weaknesses: its beam collapses onto reorderings of the same actions, and its certification reserve is too small for programs with many commuting actions. We left it unchanged so that the beyond-teacher comparison is against one fixed configuration.
The prior was trained and evaluated on corpora from one generator. The shift sets vary parameters and templates but not the generator, families or ontology, and with twenty cases each their bootstrap ranges are wide; we claim no transfer to other generators.
Certification checks executability and order relations against the compiler; it does not show that a professional would draft this way.

\section{Conclusion}
\label{sec:conclusion}

Executable patternmaking programs can be recovered from before-and-after sewing patterns and certified against a deterministic compiler.
Trained on these certified programs, a learned prior raises certified success in one fixed bounded search from 51 cases with the teacher's heuristic to 63 of 80 order-sensitive cases. It certifies more cases than every non-learned ranking on both frozen draws.
On the 50 cases that the teacher's reference configuration misses within its wall-clock budget, it certifies 34, against 21 for the heuristic.
Certified programs thus turn pattern archives, which keep the results of edits but not the steps, into supervision for search.

\section*{AI use statement}
We used generative AI tools to assist with manuscript writing and language polishing. All AI-assisted text was carefully reviewed and revised by the authors. LLM-generated code was verified and tested for correctness by the authors. We take responsibility for the final content of this work, including text, claims, and artifacts produced with the aid of generative AI.

\section*{Reproducibility statement}
Section~\ref{sec:method} specifies the drafting environment, the teacher search and its certification, the learned prior and the unified search. Section~\ref{sec:benchmark} and Appendix~\ref{app:protocol} give the corpus construction, the frozen splits, all budgets and seeds, the hardware and software, and every deviation from the protocol. Appendix~\ref{app:policy} gives the model and training details, and Appendix~\ref{app:cert} states what a certificate establishes.

\bibliography{references}

\begin{thebibliography}{42}
\providecommand{\natexlab}[1]{#1}
\providecommand{\url}[1]{\texttt{#1}}
\expandafter\ifx\csname urlstyle\endcsname\relax
  \providecommand{\doi}[1]{doi: #1}\else
  \providecommand{\doi}{doi: \begingroup \urlstyle{rm}\Url}\fi

\bibitem[Anthony et~al.(2017)Anthony, Tian, and Barber]{anthony2017thinking}
Thomas Anthony, Zheng Tian, and David Barber.
\newblock Thinking fast and slow with deep learning and tree search.
\newblock In \emph{Advances in Neural Information Processing Systems
  (NeurIPS)}, volume~30, 2017.

\bibitem[Ballesteros et~al.(2016)Ballesteros, Goldberg, Dyer, and
  Smith]{ballesteros2016training}
Miguel Ballesteros, Yoav Goldberg, Chris Dyer, and Noah~A. Smith.
\newblock Training with exploration improves a greedy stack {LSTM} parser.
\newblock In \emph{Conference on Empirical Methods in Natural Language
  Processing (EMNLP)}, pp.\  2005--2010, 2016.

\bibitem[Balog et~al.(2017)Balog, Gaunt, Brockschmidt, Nowozin, and
  Tarlow]{balog2017deepcoder}
Matej Balog, Alexander~L. Gaunt, Marc Brockschmidt, Sebastian Nowozin, and
  Daniel Tarlow.
\newblock {DeepCoder}: Learning to write programs.
\newblock In \emph{International Conference on Learning Representations
  (ICLR)}, 2017.

\bibitem[Bian et~al.(2025)Bian, Xu, Xiu, Grigorev, Liu, Lu, Black, and
  Feng]{bian2025chatgarment}
Siyuan Bian, Chenghao Xu, Yuliang Xiu, Artur Grigorev, Zhen Liu, Cewu Lu,
  Michael~J. Black, and Yao Feng.
\newblock {ChatGarment}: Garment estimation, generation and editing via large
  {Language} models.
\newblock In \emph{IEEE/CVF Conference on Computer Vision and Pattern
  Recognition (CVPR)}, 2025.

\bibitem[Chang et~al.(2015)Chang, Krishnamurthy, Agarwal, Daum{\'e}~III, and
  Langford]{chang2015learning}
Kai-Wei Chang, Akshay Krishnamurthy, Alekh Agarwal, Hal Daum{\'e}~III, and John
  Langford.
\newblock Learning to search better than your teacher.
\newblock In \emph{International Conference on Machine Learning (ICML)}, pp.\
  2058--2066, 2015.

\bibitem[Cobbe et~al.(2021)Cobbe, Kosaraju, Bavarian, Chen, Jun, Kaiser,
  Plappert, Tworek, Hilton, Nakano, Hesse, and Schulman]{cobbe2021verifiers}
Karl Cobbe, Vineet Kosaraju, Mohammad Bavarian, Mark Chen, Heewoo Jun, Lukasz
  Kaiser, Matthias Plappert, Jerry Tworek, Jacob Hilton, Reiichiro Nakano,
  Christopher Hesse, and John Schulman.
\newblock Training verifiers to solve math word problems.
\newblock arXiv preprint arXiv:2110.14168, 2021.

\bibitem[Dupont et~al.(2024)Dupont, Cherenkova, Mallis, Gusev, Kacem, and
  Aouada]{dupont2024transcad}
Elona Dupont, Kseniya Cherenkova, Dimitrios Mallis, Gleb Gusev, Anis Kacem, and
  Djamila Aouada.
\newblock {TransCAD}: A hierarchical transformer for {CAD} sequence inference
  from {Point Clouds}.
\newblock In \emph{European Conference on Computer Vision (ECCV)}, 2024.

\bibitem[Eggler et~al.(2024)Eggler, Falque, Liu, Vidal-Calleja,
  Sorkine-Hornung, and Pietroni]{eggler2024garmentalteration}
Anna~Maria Eggler, Raphael Falque, Mark Liu, Teresa Vidal-Calleja, Olga
  Sorkine-Hornung, and Nico Pietroni.
\newblock Digital garment alteration.
\newblock \emph{Computer Graphics Forum (Proc. Pacific Graphics)}, 43\penalty0
  (7), 2024.

\bibitem[Ellis et~al.(2019)Ellis, Nye, Pu, Sosa, Tenenbaum, and
  Solar-Lezama]{ellis2019write}
Kevin Ellis, Maxwell Nye, Yewen Pu, Felix Sosa, Josh Tenenbaum, and Armando
  Solar-Lezama.
\newblock Write, execute, assess: Program synthesis with a {REPL}.
\newblock In \emph{Advances in Neural Information Processing Systems
  (NeurIPS)}, 2019.

\bibitem[Ellis et~al.(2021)Ellis, Wong, Nye, Sabl{\'e}-Meyer, Morales, Hewitt,
  Cary, Solar-Lezama, and Tenenbaum]{ellis2021dreamcoder}
Kevin Ellis, Catherine Wong, Maxwell Nye, Mathias Sabl{\'e}-Meyer, Lucas
  Morales, Luke Hewitt, Luc Cary, Armando Solar-Lezama, and Joshua~B.
  Tenenbaum.
\newblock {DreamCoder}: Bootstrapping inductive program synthesis with
  {Wake-Sleep} {Library Learning}.
\newblock In \emph{ACM SIGPLAN Conference on Programming {Language} Design and
  Implementation (PLDI)}, 2021.

\bibitem[Goldberg \& Nivre(2013)Goldberg and Nivre]{goldberg2013training}
Yoav Goldberg and Joakim Nivre.
\newblock Training deterministic parsers with non-deterministic oracles.
\newblock \emph{Transactions of the Association for Computational Linguistics},
  1:\penalty0 403--414, 2013.

\bibitem[He et~al.(2024)He, Yao, Zhang, Yu, Liu, and Xu]{he2024dresscode}
Kai He, Kaixin Yao, Qixuan Zhang, Jingyi Yu, Lingjie Liu, and Lan Xu.
\newblock {DressCode}: Autoregressively sewing and generating garments from
  text guidance.
\newblock \emph{ACM Transactions on Graphics (Proc. {SIGGRAPH})}, 43\penalty0
  (4), 2024.

\bibitem[Jones et~al.(2022)Jones, Walke, and Ritchie]{jones2022plad}
R.~Kenny Jones, Homer Walke, and Daniel Ritchie.
\newblock {PLAD}: Learning to infer shape programs with pseudo-labels and
  approximate distributions.
\newblock In \emph{IEEE/CVF Conference on Computer Vision and Pattern
  Recognition (CVPR)}, 2022.

\bibitem[Jones et~al.(2024)Jones, Zhang, Ganeshan, and
  Ritchie]{jones2024vpiedit}
R.~Kenny Jones, Renhao Zhang, Aditya Ganeshan, and Daniel Ritchie.
\newblock Learning to edit visual programs with self-supervision.
\newblock In \emph{Advances in Neural Information Processing Systems
  (NeurIPS)}, volume~37, 2024.

\bibitem[Kalyan et~al.(2018)Kalyan, Mohta, Polozov, Batra, Jain, and
  Gulwani]{kalyan2018neural}
Ashwin Kalyan, Abhishek Mohta, Oleksandr Polozov, Dhruv Batra, Prateek Jain,
  and Sumit Gulwani.
\newblock Neural-guided deductive search for real-time program synthesis from
  examples.
\newblock In \emph{International Conference on Learning Representations
  (ICLR)}, 2018.

\bibitem[Khan et~al.(2024{\natexlab{a}})Khan, Dupont, Ali, Cherenkova, Kacem,
  and Aouada]{khan2024cadsignet}
Mohammad~Sadil Khan, Elona Dupont, Sk~Aziz Ali, Kseniya Cherenkova, Anis Kacem,
  and Djamila Aouada.
\newblock {CAD-SIGNet}: {CAD} {Language} inference from {Point Clouds} using
  layer-wise {Sketch Instance Guided Attention}.
\newblock In \emph{IEEE/CVF Conference on Computer Vision and Pattern
  Recognition (CVPR)}, 2024{\natexlab{a}}.

\bibitem[Khan et~al.(2024{\natexlab{b}})Khan, Sinha, Sheikh, Stricker, Ali, and
  Afzal]{khan2024text2cad}
Mohammad~Sadil Khan, Sankalp Sinha, Talha~Uddin Sheikh, Didier Stricker,
  Sk~Aziz Ali, and Muhammad~Zeshan Afzal.
\newblock {Text2CAD}: Generating sequential {CAD} designs from
  beginner-to-expert level text prompts.
\newblock In \emph{Advances in Neural Information Processing Systems
  (NeurIPS)}, 2024{\natexlab{b}}.

\bibitem[Kolodiazhnyi et~al.(2026)Kolodiazhnyi, Tarasov, Zhemchuzhnikov,
  Nikulin, Zisman, Vorontsova, Konushin, Kurenkov, and
  Rukhovich]{kolodiazhnyi2026cadrille}
Maksim Kolodiazhnyi, Denis Tarasov, Dmitrii Zhemchuzhnikov, Alexander Nikulin,
  Ilya Zisman, Anna Vorontsova, Anton Konushin, Vladislav Kurenkov, and Danila
  Rukhovich.
\newblock cadrille: Multi-modal {CAD} reconstruction with reinforcement
  learning.
\newblock In \emph{International Conference on Learning Representations
  (ICLR)}, 2026.

\bibitem[Korosteleva \& Sorkine-Hornung(2023)Korosteleva and
  Sorkine-Hornung]{korosteleva2023garmentcode}
Maria Korosteleva and Olga Sorkine-Hornung.
\newblock {GarmentCode}: Programming parametric sewing patterns.
\newblock \emph{ACM Transactions on Graphics (Proc. {SIGGRAPH} Asia)},
  42\penalty0 (6), 2023.

\bibitem[Korosteleva et~al.(2024)Korosteleva, Kesdogan, Kemper, Wenninger,
  Koller, Zhang, Botsch, and Sorkine-Hornung]{korosteleva2024garmentcodedata}
Maria Korosteleva, Timur~Levent Kesdogan, Fabian Kemper, Stephan Wenninger,
  Jasmin Koller, Yuhan Zhang, Mario Botsch, and Olga Sorkine-Hornung.
\newblock {GarmentCodeData}: A dataset of {3D} {Made-to-Measure} garments with
  sewing patterns.
\newblock In \emph{European Conference on Computer Vision (ECCV)}, 2024.

\bibitem[Li et~al.(2023)Li, Guo, Zhang, and Yan]{li2023secadnet}
Pu~Li, Jianwei Guo, Xiaopeng Zhang, and Dong-Ming Yan.
\newblock {SECAD-Net}: Self-supervised {CAD} reconstruction by learning
  {Sketch-Extrude} operations.
\newblock In \emph{IEEE/CVF Conference on Computer Vision and Pattern
  Recognition (CVPR)}, 2023.

\bibitem[Li et~al.(2024)Li, Guo, Li, Benes, and Yan]{li2024sfmcad}
Pu~Li, Jianwei Guo, Huibin Li, Bedrich Benes, and Dong-Ming Yan.
\newblock {SfmCAD}: Unsupervised {CAD} reconstruction by learning sketch-based
  feature modeling operations.
\newblock In \emph{IEEE/CVF Conference on Computer Vision and Pattern
  Recognition (CVPR)}, 2024.

\bibitem[Liu et~al.(2023)Liu, Xu, Lin, Liang, and Yan]{liu2023sewformer}
Lijuan Liu, Xiangyu Xu, Zhijie Lin, Jiabin Liang, and Shuicheng Yan.
\newblock Towards garment sewing pattern reconstruction from a single image.
\newblock \emph{ACM Transactions on Graphics (Proc. {SIGGRAPH} Asia)},
  42\penalty0 (6), 2023.

\bibitem[Mallis et~al.(2025)Mallis, Karadeniz, Cavada, Rukhovich,
  Foteinopoulou, Cherenkova, Kacem, and Aouada]{mallis2025cadassistant}
Dimitrios Mallis, Ahmet~Serdar Karadeniz, Sebastian Cavada, Danila Rukhovich,
  Niki~Maria Foteinopoulou, Kseniya Cherenkova, Anis Kacem, and Djamila Aouada.
\newblock {CAD-Assistant}: Tool-augmented {VLLMs} as generic {CAD} task
  solvers.
\newblock In \emph{IEEE/CVF International Conference on Computer Vision
  (ICCV)}, 2025.

\bibitem[Mazurkiewicz(1987)]{mazurkiewicz1987trace}
Antoni Mazurkiewicz.
\newblock Trace theory.
\newblock In \emph{Petri Nets: Applications and Relationships to Other Models
  of Concurrency (Advances in Petri Nets 1986, Part II)}, volume 255 of
  \emph{Lecture Notes in Computer Science}, pp.\  278--324. Springer, 1987.

\bibitem[Muise et~al.(2016)Muise, Beck, and McIlraith]{muise2016optimal}
Christian Muise, J.~Christopher Beck, and Sheila~A. McIlraith.
\newblock Optimal partial-order plan relaxation via {MaxSAT}.
\newblock \emph{Journal of Artificial Intelligence Research}, 57:\penalty0
  113--149, 2016.

\bibitem[Nakayama et~al.(2025)Nakayama, Ackermann, Kesdogan, Zheng,
  Korosteleva, Sorkine-Hornung, Guibas, Yang, and
  Wetzstein]{nakayama2025aipparel}
Kiyohiro Nakayama, Jan Ackermann, Timur~Levent Kesdogan, Yang Zheng, Maria
  Korosteleva, Olga Sorkine-Hornung, Leonidas Guibas, Guandao Yang, and Gordon
  Wetzstein.
\newblock {AIpparel}: A multimodal foundation model for digital garments.
\newblock In \emph{IEEE/CVF Conference on Computer Vision and Pattern
  Recognition (CVPR)}, 2025.

\bibitem[Nosek et~al.(2018)Nosek, Ebersole, DeHaven, and
  Mellor]{nosek2018preregistration}
Brian~A. Nosek, Charles~R. Ebersole, Alexander~C. DeHaven, and David~T. Mellor.
\newblock The preregistration revolution.
\newblock \emph{Proceedings of the National Academy of Sciences}, 115\penalty0
  (11):\penalty0 2600--2606, 2018.

\bibitem[Ren et~al.(2022)Ren, Zheng, Cai, Li, and Zhang]{ren2022extrudenet}
Daxuan Ren, Jianmin Zheng, Jianfei Cai, Jiatong Li, and Junzhe Zhang.
\newblock {ExtrudeNet}: Unsupervised inverse {Sketch-and-Extrude} for shape
  parsing.
\newblock In \emph{European Conference on Computer Vision (ECCV)}, 2022.

\bibitem[Rukhovich et~al.(2025)Rukhovich, Dupont, Mallis, Cherenkova, Kacem,
  and Aouada]{rukhovich2025cadrecode}
Danila Rukhovich, Elona Dupont, Dimitrios Mallis, Kseniya Cherenkova, Anis
  Kacem, and Djamila Aouada.
\newblock {CAD-Recode}: Reverse engineering {CAD} code from {Point Clouds}.
\newblock In \emph{IEEE/CVF International Conference on Computer Vision
  (ICCV)}, 2025.

\bibitem[{Seamly2D contributors}(2026)]{seamly2d}
{Seamly2D contributors}.
\newblock {Seamly2D}: Open-source pattern design software.
\newblock \url{https://seamly.io}, 2026.

\bibitem[Shi et~al.(2022)Shi, Dai, Ellis, and Sutton]{shi2022crossbeam}
Kensen Shi, Hanjun Dai, Kevin Ellis, and Charles Sutton.
\newblock {CrossBeam}: Learning to search in bottom-up program synthesis.
\newblock In \emph{International Conference on Learning Representations
  (ICLR)}, 2022.

\bibitem[Si \& Zhang(2026)Si and Zhang]{si2026checked}
Yuan Si and Jialu Zhang.
\newblock Checked program recovery from execution video: A sound oracle for
  untrusted generators.
\newblock \emph{arXiv preprint arXiv:2607.00635}, 2026.

\bibitem[Silver et~al.(2017)Silver, Schrittwieser, Simonyan, Antonoglou, Huang,
  Guez, Hubert, Baker, Lai, Bolton, et~al.]{silver2017alphago}
David Silver, Julian Schrittwieser, Karen Simonyan, Ioannis Antonoglou, Aja
  Huang, Arthur Guez, Thomas Hubert, Lucas Baker, Matthew Lai, Adrian Bolton,
  et~al.
\newblock Mastering the game of {Go} without human knowledge.
\newblock \emph{Nature}, 550\penalty0 (7676):\penalty0 354--359, 2017.

\bibitem[Singh et~al.(2024)Singh, Co-Reyes, Agarwal, et~al.]{singh2024beyond}
Avi Singh, John~D. Co-Reyes, Rishabh Agarwal, et~al.
\newblock Beyond human data: Scaling self-training for problem-solving with
  language models.
\newblock \emph{Transactions on Machine Learning Research}, 2024.

\bibitem[Sun et~al.(2026)Sun, Wang, Liu, Kong, He, He, and
  Wang]{sun2026tailorcopilot}
Yuexin Sun, Zhaohui Wang, Ruiyang Liu, Demian Kong, Qian He, Gaofeng He, and
  Huamin Wang.
\newblock {TailorCoPilot}: Enabling agentic pattern making with
  version-controlled state tracking.
\newblock \emph{arXiv preprint arXiv:2608.25462}, 2026.
\newblock To appear at ACM UIST 2026.

\bibitem[Vaswani et~al.(2017)Vaswani, Shazeer, Parmar, Uszkoreit, Jones, Gomez,
  Kaiser, and Polosukhin]{vaswani2017attention}
Ashish Vaswani, Noam Shazeer, Niki Parmar, Jakob Uszkoreit, Llion Jones,
  Aidan~N. Gomez, {\L}ukasz Kaiser, and Illia Polosukhin.
\newblock Attention is all you need.
\newblock In \emph{Advances in Neural Information Processing Systems
  (NeurIPS)}, 2017.

\bibitem[Weng et~al.(2026)Weng, Chen, Li, Qin, Guo, Hao, and
  Han]{weng2026garmentgpt}
Fangsheng Weng, Junhao Chen, Xiang Li, Jie Qin, Hanzhong Guo, Shaochun Hao, and
  Xiaoguang Han.
\newblock {GarmentGPT}: Compositional garment pattern generation via discrete
  latent tokenization.
\newblock In \emph{International Conference on Learning Representations
  (ICLR)}, 2026.

\bibitem[Wu et~al.(2021)Wu, Xiao, and Zheng]{wu2021deepcad}
Rundi Wu, Chang Xiao, and Changxi Zheng.
\newblock {DeepCAD}: A deep generative network for computer-aided design
  models.
\newblock In \emph{IEEE/CVF International Conference on Computer Vision
  (ICCV)}, 2021.

\bibitem[Xu et~al.(2022)Xu, Willis, Lambourne, Cheng, Jayaraman, and
  Furukawa]{xu2022skexgen}
Xiang Xu, Karl D.~D. Willis, Joseph~G. Lambourne, Chin-Yi Cheng, Pradeep~Kumar
  Jayaraman, and Yasutaka Furukawa.
\newblock {SkexGen}: Autoregressive generation of {CAD} construction sequences
  with disentangled codebooks.
\newblock In \emph{International Conference on Machine Learning (ICML)}, 2022.

\bibitem[Xu et~al.(2021)Xu, Peng, Cheng, Willis, and Ritchie]{xu2021zonegraphs}
Xianghao Xu, Wenzhe Peng, Chin-Yi Cheng, Karl D.~D. Willis, and Daniel Ritchie.
\newblock Inferring {CAD} modeling sequences using {Zone Graphs}.
\newblock In \emph{IEEE/CVF Conference on Computer Vision and Pattern
  Recognition (CVPR)}, 2021.

\bibitem[Zhou et~al.(2025)Zhou, Liu, Liu, He, Li, Jin, and
  Wang]{zhou2025design2garmentcode}
Feng Zhou, Ruiyang Liu, Chen Liu, Gaofeng He, Yong-Lu Li, Xiaogang Jin, and
  Huamin Wang.
\newblock {Design2GarmentCode}: Turning design concepts to tangible garments
  through program synthesis.
\newblock In \emph{IEEE/CVF Conference on Computer Vision and Pattern
  Recognition (CVPR)}, 2025.

\end{thebibliography}
\bibliographystyle{iclr2027_conference}

\clearpage
\appendix

\section*{Supplementary material}
This supplementary material accompanies the paper ``Learning to Search for Executable Patternmaking from Certified Programs''. Sections, figures, tables and equations are numbered on from the main paper, whose reference list also covers the work cited here.

\section{Protocol details}
\label{app:protocol}

\paragraph{Search configuration (Tier C).}
Beam width 96, top-$k$ 8 children per prefix, per-provider proposal cap, maximum depth 8, execution cap 20,000, wall-clock search budget 1,350 s with a 1,800 s emergency stop, 512 contour samples for evaluation and 128 for the lookahead landing check, certification reserve 25\% of the wall budget. The lookahead follows each proposed detach with the admissible re-attach variants and keeps the pair only if some variant lands the swapped panels within the tight thresholds. The certification reserve stops the search at 75\% of the wall-clock budget once a tight terminal exists, so that certification can complete.
Tier A and B budgets are 225/300 s and 450/600 s of wall clock, and 2,500 and 7,500 executions.
Tie-break order A is the main run; order B repeats Tier C with the alternative deterministic order and certifies exactly the same 57 of 80 cases (17 budget-exhausted and 6 timeouts against 18 and 5 for order A, with one case moving from budget-exhausted to timeout), so the Tier-C failures are properties of the cases, not of the tie-break.

\paragraph{Hardware and software.}
The teacher's runs on \fullb{} used a cloud virtual machine with 8 vCPUs (AMD EPYC 9T95) and 30 GB of RAM.
All runs on the frozen evaluation sets, the training of the prior and the development-stage rollouts used one server with two 32-core Intel Xeon Platinum 8350C CPUs, 1 TB of RAM and eight NVIDIA A800 GPUs with 80 GB each.
Every search runs single-threaded in its own process; the prior runs on one GPU, and H, R and T use none.
On both machines a Tier-C teacher search runs at about one engine execution per second, so a wall-clock budget buys a similar number of executions on either.
The compiler, the searches and the certifier use only the Python standard library (CPython 3.10 on Ubuntu 22.04); the prior uses PyTorch 2.5.1 with CUDA 12.1.

\paragraph{Protocol versions.}
v9 is the baseline hierarchical search (Tier A results).
v10 added the template-swap lookahead; v10.1 the landing admission; v10.2 and v10.3 a cheaper landing check and batched evaluation; v10.4 the certification reserve; v10.4.1 corrected a defect in v10.4 in which the reserve was also applied to the correspondence used by the DAG endpoint assessment, which had produced spurious ``DAG incomplete'' verdicts.
Each version has a protocol written before its runs and a run record; the held-out Tier-C run used v10.4.1.

\paragraph{Deviations logged.}
A read-only trace-extraction job (11 min, low priority) ran on the same machine during one Tier-C case; the case was re-run on an idle machine with identical execution count.
Five cases whose result serialisation exceeded the emergency stop are counted as timeouts in the table fixed before the runs: a controlled re-run with the watchdog doubled and the search deadline unchanged showed that all five had finished searching (beam collapse) and were killed 19--41 s past the hard stop while serialising a result file of roughly a hundred megabytes. The independent re-run of one failed case on an idle machine reproduced the original execution count exactly, ruling out interference from a concurrent job.
The Tier-B re-run under v10.4.1 and the order-B Tier-C run were moved from before the training of the prior to in parallel with it.
On the frozen main test, the teacher's reference configuration ran in parallel with the comparison runs on one shared machine, first in two and then in five concurrent processes; one case that ended in a runner error was re-run from a fresh directory and certified.
Baseline ordering fix: in the first runs of R and T, children were keyed by their negated compile rank within each prefix, so that at beam 4 the beam kept the lowest-ranked children first. After this one-line correction of the search script, R and T at beam 4 were re-run on all four frozen test sets on 2026-09-19 with every other setting unchanged. Control runs of T and of one R seed at beam 1, where the key has no effect, reproduced the original per-case outcomes exactly; H and P use a different key and are unaffected. All R and T numbers in the paper come from the corrected runs.

\paragraph{Frozen evaluation set.}
Generated on 2026-09-12 with the constructor of \fullb{} under a new corpus label; the digests of its configuration and case manifest were recorded before use.
The constructor was modified only to accept the new label; the patch is recorded with the artifact.
The output root was made read-only before any run, and the splits (24 calibration cases, main test 80, easy reference 20) were written from the public manifest by a script that also de-duplicates against \fullb{} by target state and artifact hash.
Calibration on the 24 calibration cases (eight per tier, all Tier-C ones at depth 6) fixed the protocol on 2026-09-12 before any run on the main test: because the search path does not depend on the budget, each condition is run once at 20\% of the tier's execution cap (1,500 for Tier B, 4,000 for Tier C) and success at any smaller budget is read off the execution count of the first tight state; the reported budget points are 0.5, 1, 2, 5, 10 and 20\% of the cap; the primary endpoint is certified success at 5\% with beam width 4, beam width 1 is secondary; R uses seeds 1--3, P-DAG training seeds 1--2. On these cases the results at 5\% and 20\% were identical for all fifteen configurations, beam width 4 solved the two depth-6 cases that beam width 1 lost for the learned prior, and the teacher heuristic used greedily solved all eight Tier-C calibration cases with a median of 24 executions; the template lookup is built from all 1,510 certified programs, and the learned prior is trained on the training part of the same programs (Sec.~\ref{sec:student-results}).

\paragraph{Post-hoc robustness experiments.}
Three experiments were added after the main results had been read; each was written down with its interpretation table and frozen (file digests recorded) before its first run, and none changed the prior or any setting of the main protocol.
Beam de-duplication: prefixes whose full state hash and set of used action identifiers coincide are merged and the first-ranked one kept; this merges only reorderings of the same actions, whose remaining candidates are identical, and is applied to the heuristic and the prior under the same rule. In the recorded traces the heuristic's beam held a median of 12 merged states per case against 2 for the prior, so de-duplication mostly changes the heuristic's search, and the freed slots go to children of worse-ranked distinct prefixes that the plateaued metric ranks above the right branch; a beam of distinct states pays only when the ranking signal can tell them apart.
Distribution shift: the generator copy accepts a mode switch that either replaces the Tier-C parameter value sets or replaces the Tier-C templates; two 300-case corpora were built under their own corpus labels and frozen read-only, and the twenty test cases per corpus are the held-out positions 2, 5, 11, 15 and 21 of each family, fixed before the build. The parameter-range shift moves dart amounts and depths, slash amounts, transfer positions and depths, length changes and ease amounts to unseen values inside the compilable range. The template shift places ease and a primary-panel slash before the swap, removes the length change at depth 6 and adds a style line at depth 7. Each set has five cases per family with depths 6/7/8 in the proportion 8/8/4, and no target coincides with any earlier corpus.
The first template and parameter definitions were rejected by the constructor's admission rules (a two-dart commuting head, a structural final action, and a compilable ease amount are required); the revisions were made before any ranker ran and are logged with the frozen protocol.
Replication: the second draw was generated on 2026-09-13 with the same generator, templates and parameter value sets under another corpus label, with its configuration and manifest digests recorded, and made read-only; its test set is the 80 held-out Tier-C positions under the rule of the main test (depths 6/7/8: 28/28/24, 20 per family), with no target shared with any earlier corpus. The replication protocol fixed the reading before any ranker ran: the main conclusion replicates if the paired gain over the heuristic again has a bootstrap range above zero, replicates in direction only if the gain stays positive with a range that includes zero, and is rewritten if the sign reverses.

\section{Full tables of the frozen comparison}
\label{app:tables}
Tables~\ref{tab:main} and~\ref{tab:shift} give every condition of the comparison summarised in Figs.~\ref{fig:policy} and~\ref{fig:robustness}, with search and certification costs. Table~\ref{tab:paired} gives the paired gains, and Table~\ref{tab:cells} splits the two draws by family and target depth.

\begin{table}[htbp]
\centering
\small
\caption{Main comparison on the 80 Tier-C main-test cases (frozen evaluation set). All conditions use the same candidates, beam width, public-metric stopping rule, certifier and execution budget (5\% of the tier cap, 1,000 executions); only the ranking signal differs. R is the mean over three seeds and P-DAG over two training seeds. Search executions: median engine executions to the first tight state over successful cases; certification executions: median over certified cases; wall clock: median seconds per case including certification. Certified at 1\% / 2\%: the same runs truncated at 200 / 400 executions.}
\label{tab:main}
\setlength{\tabcolsep}{4pt}
\begin{tabular}{llcccccccc}
\toprule
Ranker & beam & tight & certified & cert.\ at 1\% & at 2\% & search exec. & cert.\ exec. & wall (s) \\
\midrule
H: teacher heuristic & 1 & 52 & 51 & 49 & 51 & 103 & 43 & 117 \\
R: random ($\times$3) & 1 & 18 & 17.7 & 17.7 & 17.7 & 15 & 51 & 14 \\
T: template lookup & 1 & 36 & 27 & 27 & 27 & 14 & 40 & 17 \\
P: prior, linear & 1 & 41 & 35 & 35 & 35 & 14 & 43 & 17 \\
P: prior, DAG ($\times$2) & 1 & 41 & 35 & 35 & 35 & 14 & 43 & 17 \\
\midrule
H: teacher heuristic & 4 & 52 & 51 & 4 & 29 & 351 & 43 & 349 \\
R: random ($\times$3) & 4 & 42 & 37.7 & 27.3 & 37.7 & 178 & 43 & 61 \\
T: template lookup & 4 & 59 & 53 & 39 & 53 & 176 & 40 & 60 \\
P: prior, linear & 4 & 69 & 61 & 54 & 61 & 176 & 40 & 61 \\
P: prior, DAG ($\times$2) & 4 & 71 & 63 & 54 & 63 & 180 & 40 & 61 \\
\bottomrule
\end{tabular}
\end{table}

\begin{table}[htbp]
\centering
\small
\caption{Robustness of the main comparison: certified cases at the 5\% budget on the main test, on an independent replication draw from the same generator (80 further Tier-C cases, protocol frozen before the run), and under two distribution shifts (20 cases each). Same conditions as Table~\ref{tab:main}; the prior is not retrained. Table~\ref{tab:paired} gives the paired gains.}
\label{tab:shift}
\footnotesize
\setlength{\tabcolsep}{3pt}
\resizebox{\linewidth}{!}{\begin{tabular}{lcccccccc}
\toprule
 & \multicolumn{2}{c}{main test (80)} & \multicolumn{2}{c}{replication (80)} & \multicolumn{2}{c}{parameter shift (20)} & \multicolumn{2}{c}{template shift (20)} \\
Ranker & b.\,1 & b.\,4 & b.\,1 & b.\,4 & b.\,1 & b.\,4 & b.\,1 & b.\,4 \\
\midrule
H: teacher heuristic & 51 & 51 & 52 & 52 & 11 & 11 & 10 & 10 \\
R: random ($\times$3) & 17.7 & 37.7 & 16.3 & 37.3 & 1.7 & 9.7 & 3.3 & 9.3 \\
T: template lookup & 27 & 53 & 33 & 51 & 7 & 9 & 6 & 10 \\
P: prior, linear & 35 & 61 & 47 & 64 & 10 & 11 & 10 & 13 \\
P: prior, DAG ($\times$2) & 35 & 63 & 46 & 64 & 10 & 13 & 9 & 14 \\
\bottomrule
\end{tabular}}
\end{table}

\begin{table}[htbp]
\centering
\small
\caption{Paired gains in certified success at beam 4 within 1,000 search executions, in cases and in percentage points of each set. P is the prior with DAG supervision averaged over its two seeds, except where linear imitation is named. The range spans the 2.5th to 97.5th percentiles of 2,000 bootstrap resamples of the cases.}
\label{tab:paired}
\setlength{\tabcolsep}{4pt}
\begin{tabular}{llrrr}
\toprule
Set (cases) & Comparison & Gain (cases) & Gain (points) & Range (points) \\
\midrule
Main test (80) & P over H & 12 & 15.0 & 3.8 to 27.5 \\
 & P over T & 10 & 12.5 & 3.1 to 21.9 \\
 & P over R & 25.3 & 31.7 & 21.0 to 42.3 \\
 & P with linear imitation over H & 10 & 12.5 & 1.3 to 25.0 \\
\addlinespace
Replication (80) & P over H & 12 & 15.0 & 6.3 to 25.0 \\
 & P over T & 13 & 16.3 & 8.8 to 25.0 \\
 & P over R & 26.7 & 33.3 & 25.4 to 41.7 \\
\addlinespace
Both draws (160) & P over H & 24 & 15.0 & 7.5 to 22.8 \\
\addlinespace
Parameter shift (20) & P over H & 2 & 10.0 & $-7.5$ to 27.5 \\
 & P over T & 4 & 20.0 & 7.5 to 35.0 \\
 & P over R & 3.3 & 16.7 & $-3.3$ to 35.0 \\
\addlinespace
Template shift (20) & P over H & 4 & 20.0 & 0.0 to 45.0 \\
 & P over T & 4 & 20.0 & 5.0 to 40.0 \\
 & P over R & 4.7 & 23.3 & 8.3 to 40.0 \\
\addlinespace
Beam de-duplication (80) & P merged over P & 3 & 3.8 & 1.3 to 6.9 \\
 & H merged over H & $-8$ & $-10.0$ & $-18.8$ to $-1.3$ \\
 & P over H, both merged & 23 & 28.8 & 13.8 to 42.5 \\
\bottomrule
\end{tabular}
\end{table}

\begin{table}[htbp]
\centering
\small
\caption{Certified cases by family and target depth at beam 4 within 1,000 search executions, as P-DAG (mean of two seeds) / T / H. Depth-6 and depth-7 cells have seven cases and depth-8 cells six. Resampling the twelve family-by-depth cells instead of the cases widens the bootstrap range of the gain of 15 points over H to $-10$ to 40 points on the main test and to $-4$ to 37 points on the replication draw.}
\label{tab:cells}
\resizebox{\linewidth}{!}{\begin{tabular}{lcccc cccc}
\toprule
 & \multicolumn{4}{c}{main test} & \multicolumn{4}{c}{replication} \\
\cmidrule(lr){2-5}\cmidrule(lr){6-9}
Family & depth 6 & depth 7 & depth 8 & all (20) & depth 6 & depth 7 & depth 8 & all (20) \\
\midrule
tee   & 7 / 3 / 7 & 7 / 6 / 1 & 2.5 / 3 / 0 & 16.5 / 12 / 8 & 7 / 6 / 7 & 7 / 5 / 1 & 2 / 1 / 0 & 16 / 12 / 8 \\
skirt & 7 / 7 / 7 & 7 / 7 / 7 & 6 / 1 / 6 & 20 / 15 / 20 & 7 / 7 / 7 & 7 / 7 / 7 & 6 / 1 / 6 & 20 / 15 / 20 \\
pants & 7 / 7 / 7 & 4 / 5 / 7 & 0 / 0 / 4 & 11 / 12 / 18 & 7 / 7 / 7 & 7 / 5 / 7 & 0 / 0 / 3 & 14 / 12 / 17 \\
dress & 7 / 7 / 4 & 7 / 6 / 1 & 1.5 / 1 / 0 & 15.5 / 14 / 5 & 7 / 6 / 6 & 7 / 6 / 1 & 0 / 0 / 0 & 14 / 12 / 7 \\
\midrule
all   & 28 / 24 / 25 & 25 / 24 / 16 & 10 / 5 / 10 & 63 / 53 / 51 & 28 / 26 / 27 & 28 / 23 / 16 & 8 / 2 / 9 & 64 / 51 / 52 \\
\bottomrule
\end{tabular}}
\end{table}

\section{Development-stage results on \fullb{}}
\label{app:devresults}

The results in this appendix were obtained on \fullb{} before the frozen evaluation set existed, under the earlier rollout protocol (model-triggered STOP allowed, budget 5\% of the teacher cap unless stated).
They guided the protocol of Sec.~\ref{sec:unified} and are reported for completeness; the main text makes no claim from them. In these rollouts the prior acts directly as a policy that picks the next action and may stop by itself, so this appendix calls it the policy; its random held-out split is the validation split of Sec.~\ref{sec:student-results}.

\paragraph{Rollouts, controls and certification.}

\begin{table}[h]
\centering
\small
\caption{Rollouts on the random held-out split with 5\% of the teacher's per-tier execution cap. Success is the public tight criterion. Executions are the median and maximum engine executions of successful episodes. Certified: the successful programs re-verified with the teacher's full certification.}
\label{tab:rollout-dev}
\begin{tabular}{lccccccc}
\toprule
Ranker & A (15) & B (10) & C (11) & all (36) & exec. med / max & certified \\
\midrule
Policy, DAG supervision, greedy & 15 & 10 & 10 & 35 & 14 / 30 & 35/35 \\
Policy, DAG supervision, beam 4 & 15 & 10 & 11 & 36 & 56 / 278 & -- \\
Policy, linear imitation, greedy & 15 & 10 & 10 & 35 & 14 / 30 & -- \\
Random ranker (control) & 14 & 10 & 2 & 26 & 10 / 30 & -- \\
\bottomrule
\end{tabular}
\end{table}

\begin{figure}[h]
\centering
\includegraphics[width=\linewidth]{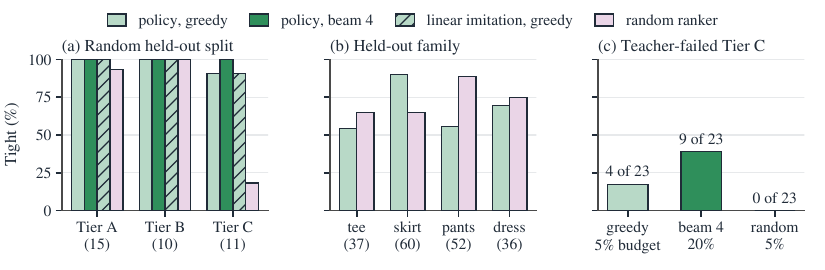}
\caption{Policy rollouts. (a) Random held-out split: the policy matters only on Tier C, where a random ranker over the same legal candidates fails. (b) Held-out family: the policy is below the random control on three of four families. (c) The 23 Tier-C cases the teacher could not solve: the policy-guided beam of four certifies seven of them with about 300 executions each.}
\label{fig:policy-dev}
\end{figure}

Table~\ref{tab:rollout-dev} and Fig.~\ref{fig:policy-dev}a report closed-loop rollouts.
With greedy decoding the policy reaches a tight endpoint on 35 of 36 held-out cases using a median of 14 engine executions per case; the teacher's median on the same Tier-C cases is above 1,400.
All 35 programs pass the teacher's three-stage certification unchanged.
With a beam of four the last Tier-C case is also solved.
The random-ranker control is the most informative row: it succeeds on 24 of the 25 Tier-A and Tier-B cases, because the environment's proposal providers already constrain the candidates so tightly that almost any consistent choice reaches the target within the depth limit, but it solves only 2 of 11 Tier-C cases against the policy's 10.
The policy's measurable contribution is therefore ordering: knowing that the sleeve swap comes after the length change and the darts.
Linear imitation matches DAG supervision here as well.

\paragraph{Policy on the cases the teacher failed.}
The 23 Tier-C cases the teacher failed are outside the training data.
Greedy rollout with the 5\% budget reaches a tight endpoint on 4 of them; a beam of four with a 20\% budget (4,000 executions) reaches tight on 9, using 279--325 executions each, and 7 of the 9 programs pass full certification (Table~\ref{tab:unsolved-dev}).
The random ranker solves none.
Four of the seven are tee variants whose eight-step programs place the swap after the length and dart actions and before an ease redistribution and a slash-and-spread, exactly the region where the teacher's beam collapsed; two are the dress cases where the teacher's lookahead had landed but lost the swapped children.
The two uncertified programs pass the order check and the independent replay but fail materialisation of one downward-closed subset with the control-certification error described in Appendix~\ref{app:cert}.
On the same 23 cases the teacher had spent 1,600--2,700 executions and its full wall clock.

\begin{table}[h]
\centering
\small
\caption{Rollouts on the 23 Tier-C cases the teacher failed (no training data from these cases) and on the generalisation splits. Family: the policy is trained on the other three families; random control uses the same candidates and budget. Composition: trained on Tiers A and B only, tested on all 57 certified Tier-C cases.}
\label{tab:unsolved-dev}
\begin{tabular}{lcc|lcc}
\toprule
Teacher-failed Tier C (23) & tight & certified & Generalisation split & policy & random \\
\midrule
Policy, greedy, 5\% budget & 4 & 4 & family tee (37; C 14) & 20 (C 0) & 24 (C 3) \\
Policy, beam 4, 20\% budget & 9 & 7 & family skirt (60; C 20) & 54 (C 16) & 39 (C 6) \\
Random ranker, 5\% budget & 0 & 0 & family pants (52; C 14) & 29 (C 0) & 46 (C 9) \\
 & & & family dress (36; C 9) & 25 (C 0) & 27 (C 0) \\
 & & & composition (57 Tier C) & 0 & 20 \\
\bottomrule
\end{tabular}
\end{table}

\paragraph{Generalisation: negative results.}
Table~\ref{tab:unsolved-dev} (right) and Fig.~\ref{fig:policy-dev}b report the held-out-family and held-out-composition splits.
When a family is held out, the policy is better than the random control only for skirts; for tees, pants and dresses it is worse, and it solves none of the held-out family's Tier-C cases, because it confidently applies the orderings of the other families where they do not hold, whereas the random ranker sometimes stumbles onto the right order.
When every case containing a swap is held out, the policy solves 0 of 57 Tier-C cases against 20 for the random ranker: it has never seen a template swap and assigns it negligible probability.
The family gate fixed before the runs, a rollout rate of at least half the random-split rate, is met on every fold; the random control is the stricter comparison and is the one used in the main protocol.

\paragraph{Teacher results by tier.}
On Tier A every held-out case is tight and 77 are certified; the three uncertified cases are tight endpoints whose order check did not complete within its allowance, the same mechanism as on Tier B below.
Adding the lookahead and the certification reserve to the search raises Tier B from 51 to 70 certified cases with no regression on any case; the nine remaining Tier-B cases are tight but uncertified because their many commuting pairs exhaust the order check's execution allowance.
On Tier C, 57 of 80 cases are certified (71\%), above the gate set for this development run; every tight case is also certified, so the certification never rejected a tight endpoint on this tier.
The 23 non-certified Tier-C cases were classified from their recorded beams.
Sixteen are beam collapse: the surviving prefixes are reorderings of the same one or two sets of actions and reach only one or two distinct states.
Three pants cases are wall-clock limited, and the remaining four show mixed causes (Table~\ref{tab:failures}).

\begin{figure}[h]
\centering
\includegraphics[width=0.5\linewidth]{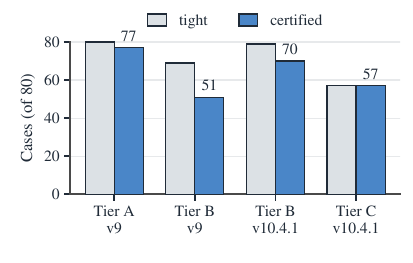}
\caption{Teacher on the held-out positions of \fullb{}: tight and certified rates per tier.}
\label{fig:teacher}
\end{figure}

\begin{table}[h]
\centering
\small
\caption{Trace-level classification of the 23 non-certified Tier-C held-out cases. Beam collapse: the beam fills with reorderings of the same actions. Distinct states is the number of different pattern states among the surviving beam prefixes at the last completed depth.}
\label{tab:failures}
\footnotesize
\begin{tabular}{lccp{4.2cm}}
\toprule
Mechanism & cases & families & evidence in the trace \\
\midrule
Beam collapse & 16 & tee 6, pants 1, dress 9 & 44--96 prefixes but 1--2 distinct states; no lookahead landing (11) or landed but displaced (5 dress) \\
Wall-clock limited & 3 & pants 3 & depth 5 of 8, 7 distinct states, best Chamfer $0.002$ \\
Mixed & 4 & pants 2, dress 2 & 3--4 states or empty frontier at deadline \\
\midrule
of which result-write timeout & 5 & tee 1, dress 4 & search finished; killed 19--41 s past the hard stop while writing the result \\
\bottomrule
\end{tabular}
\end{table}

\begin{table}[h]
\centering
\small
\caption{Teacher results on the held-out positions of \fullb{} (80 cases per tier); these are oracle-informed development results (Sec.~\ref{sec:teacher}). Tight is the public endpoint criterion; certified additionally passes recompilation, the order check and the independent replay. Tier A was run under protocol v9; Tiers B and C under v10.4.1 (Sec.~\ref{sec:teacher}). Per-family counts are certified cases out of 20.}
\label{tab:teacher}
\resizebox{\linewidth}{!}{\begin{tabular}{lcccc cccc}
\toprule
& \multicolumn{4}{c}{overall (of 80)} & \multicolumn{4}{c}{certified per family} \\
\cmidrule(lr){2-5}\cmidrule(lr){6-9}
Tier (protocol) & tight & certified & search failed & write timeout & tee & skirt & pants & dress \\
\midrule
A (v9) & 80 & 77 & 0 & 0 & -- & -- & -- & -- \\
B (v9) & 69 & 51 & 11 & 0 & -- & -- & -- & -- \\
B (v10.4.1) & 79 & 70 & 1 & 0 & 19 & 20 & 20 & 11 \\
C (v10.4.1) & 57 & 57 & 18 & 5 & 14 & 20 & 14 & 9 \\
\bottomrule
\end{tabular}}
\end{table}

\begin{table}[h]
\centering
\small
\caption{Beyond the teacher: certified cases within 1,000 executions on the 50 main-test cases that the teacher's reference configuration misses, by its failure mode (beam 4; beam 1 in parentheses). R is the mean over three seeds, P-DAG the count certified by both training seeds. The reference configuration itself certifies 30 cases (17 of 28 at depth 6, 10 of 28 at depth 7 and 3 of 24 at depth 8), with a median of 1,320 executions and 1,373 s each. On these 30 cases, P-DAG certifies 27 with both seeds, T 24 and H 30.}
\label{tab:c3}
\begin{tabular}{lccccc}
\toprule
Reference failure mode & $n$ & H & R & T & P-DAG \\
\midrule
tight, not certified & 5 & 5 & 3.7 & 4 & 5 \\
search envelope exhausted & 27 & 10 & 7.3 & 16 & 20 \\
result-write overrun after search & 18 & 6 & 3 & 9 & 9 \\
\midrule
all & 50 & 21 (21) & 14 (5.7) & 29 (10) & 34 (14) \\
\bottomrule
\end{tabular}
\end{table}

\section{Details of the learned prior}
\label{app:policy}

Training programs: the certified terminals on the 80 held-out positions per tier of \fullb{}, on which Sec.~\ref{sec:teacher-results} reports the teacher; these positions are distinct from the validation split of Sec.~\ref{sec:student-results}. The Tier-B programs come from the v9 run (51 cases), because the v10.4.1 re-run of Tier B proceeded in parallel with the training (Appendix~\ref{app:protocol}).

Tokens: global; current panels (role, component, width, height, area, vertex count, per-panel Chamfer/Hausdorff/RMS to the matched target panel, size differences, edge-semantic multi-hot); target panels; components; darts and target darts (panel, boundary, centre fraction, intake, depth); executed actions (type, anchor).
Model: 4-layer Transformer encoder, $d{=}128$, 4 heads, dropout 0.1; heads for type (10 classes), anchor (pointer over panel and component tokens conditioned on the type), numeric parameters (amount, intake, depth, centre fraction, spread, ease) and categorical slots (template variant, axis, side, pivot, boundary side).
Optimisation (under five minutes on one GPU): AdamW, learning rate $3\times10^{-4}$, weight decay 0.01, cosine schedule, batch 64, 40 epochs, gradient clipping 1.0.
Loss: Eq.~\eqref{eq:dag-loss} for the type and for the anchor given the type, smooth-$\ell_1$ on numeric parameters (millimetres scaled by 100) and cross-entropy on categorical slots, with weights 1/1/0.5/0.5.
Candidate score, used by P in the unified search of Sec.~\ref{sec:unified} and in the development-stage rollouts:
\begin{equation}
\mathrm{score}(a) = \log p(\text{type}) + \log p(\text{anchor}\mid\text{type}) + 0.5\sum_{j}\log p(\text{slot}_j) - 0.5\sum_{i}\frac{|\hat{v}_i - v_i|}{\sigma_i},
\label{eq:score}
\end{equation}
where $\text{slot}_j$ are the categorical slots of the candidate, $v_i$ its numeric parameters, $\hat{v}_i$ the values that the prior predicts for them, and $\sigma_i$ a fixed scale of 25 mm, or 0.25 for normalised parameters.
In the development-stage rollouts of Appendix~\ref{app:devresults}, STOP is chosen when its log-probability exceeds that of the best candidate's type.
Development-stage budgets: 125/375/1,000 executions for Tiers A/B/C (5\%); 4,000 for the 20\% beam-4 run on the teacher-failed cases.

\begin{table}[h]
\centering
\small
\caption{Teacher-forced next-action accuracy on the random held-out split (36 cases, 529 states). Majority-class baselines and a mean predictor for the millimetre amount are shown for calibration. Two seeds of DAG supervision differ by 0.2 points.}
\label{tab:tf}
\begin{tabular}{lcccccc}
\toprule
Supervision & type & majority & anchor & joint & STOP & amount MAE / mean pred. (mm) \\
\midrule
DAG (seed 1) & 0.919 & 0.488 & 0.990 & 0.911 & 1.000 & 4.9 / 10.9 \\
DAG (seed 2) & 0.917 & 0.488 & 0.990 & 0.909 & 1.000 & 4.9 / 10.9 \\
Linear imitation & 0.919 & 0.446 & 0.986 & 0.909 & 1.000 & 5.8 / 11.6 \\
\bottomrule
\end{tabular}
\end{table}

\section{Certification of programs found with the prior}
\label{app:cert}

Programs found with the prior are certified by the same three stages as teacher terminals: recompilation of the whole program from the base state; the pairwise order check and subset materialisation of Sec.~\ref{sec:teacher}, with the public endpoint assessment (exact structure, topology, seams, darts, tags, tight) and the requirement that the materialised full set carries a STOP control decision whose artifact hash equals the recompiled endpoint; and the oracle-isolated independent replay with its reserved execution budget.
A program is certified only if the order check and materialisation are complete and the independent replay confirms the endpoint.
The model-triggered STOP described in Appendix~\ref{app:policy} was used only in the development-stage rollouts of Appendix~\ref{app:devresults}; in the unified search of Sec.~\ref{sec:unified} an episode ends only when the public metric reports a tight state, and the same rule applies to every ranking signal.

\paragraph{What a certificate establishes.}
Pairs with a static conflict and pairs that never become adjacent keep the recovered order without an executed test, so the certified partial order is conservative and some valid orders may be excluded.
A commuting verdict is obtained by execution at one shared prefix and reused at the others, and each downward-closed subset is materialised once, along its canonical order, together with every valid next action from it.

\paragraph{Order counterfactual for the worked case.}
The certificate of the program in Fig.~\ref{fig:walkthrough} keeps the precedence of the length adjustment over the attachment as an implied one, without an executed test, so we compiled the same six actions in two other orders.
With the swap moved before the length adjustment, the attached sleeves measure 36.0 by 13.2\,cm instead of 37.5 by 13.8\,cm; structure, darts and topology still match the target, and the contour distances (Chamfer 0.0052, Hausdorff 0.0204, RMS 0.0107) exceed the tight thresholds.
Detaching before and attaching after the length adjustment reproduces the certified artifact hash exactly.
The attachment therefore has to follow the length adjustment to reproduce this target.

\paragraph{Tight endpoints that fail certification.}
Every tight endpoint of Sec.~\ref{sec:main-results} that fails certification and that we examined fails in the same way: the program passes the pairwise order check and the independent replay, but the certifier refuses to materialise it because a proper downward-closed subset, the program without its ease-redistribution step, is already tight, so the full program contains an action the endpoint metric cannot justify; the certifier's minimality rule then rejects the STOP at the full set.
These are executable programs that reach the target with one redundant step, and they count as uncertified under the protocol.

\section{Exporting certified trajectories to a drafting tool}
\label{app:seamly}

\begin{figure}[h]
\centering
\includegraphics[width=0.48\linewidth]{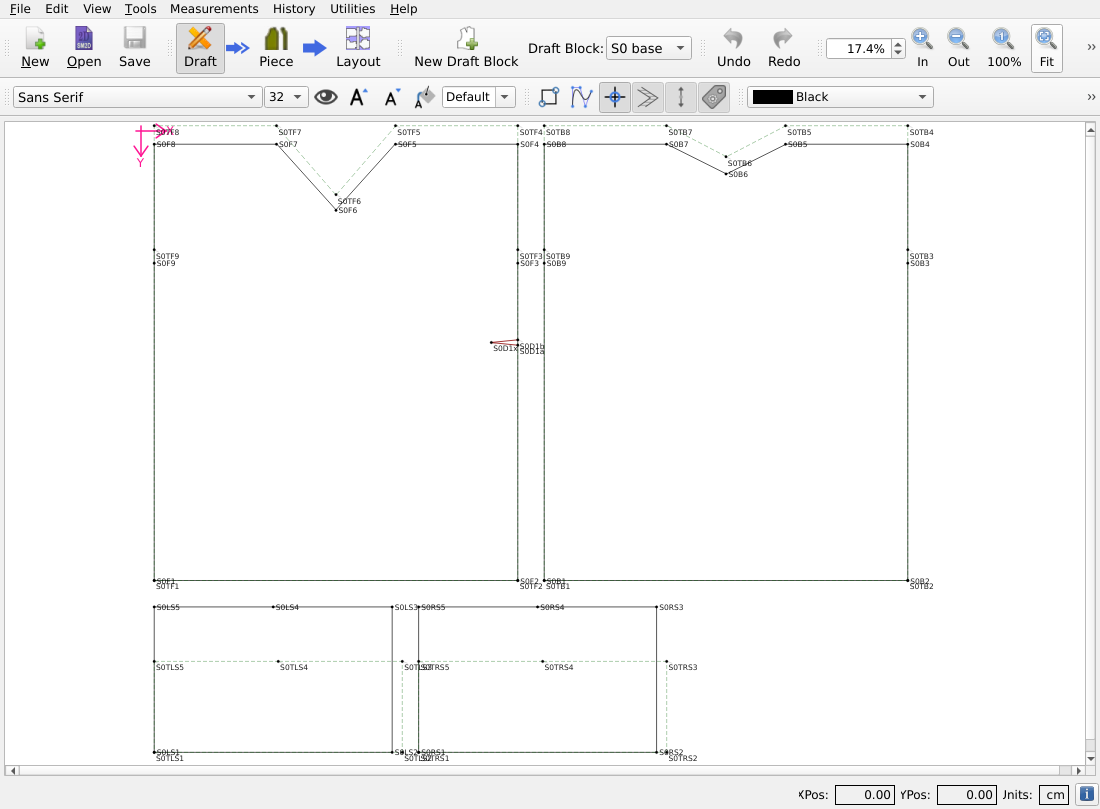}\hfill
\includegraphics[width=0.48\linewidth]{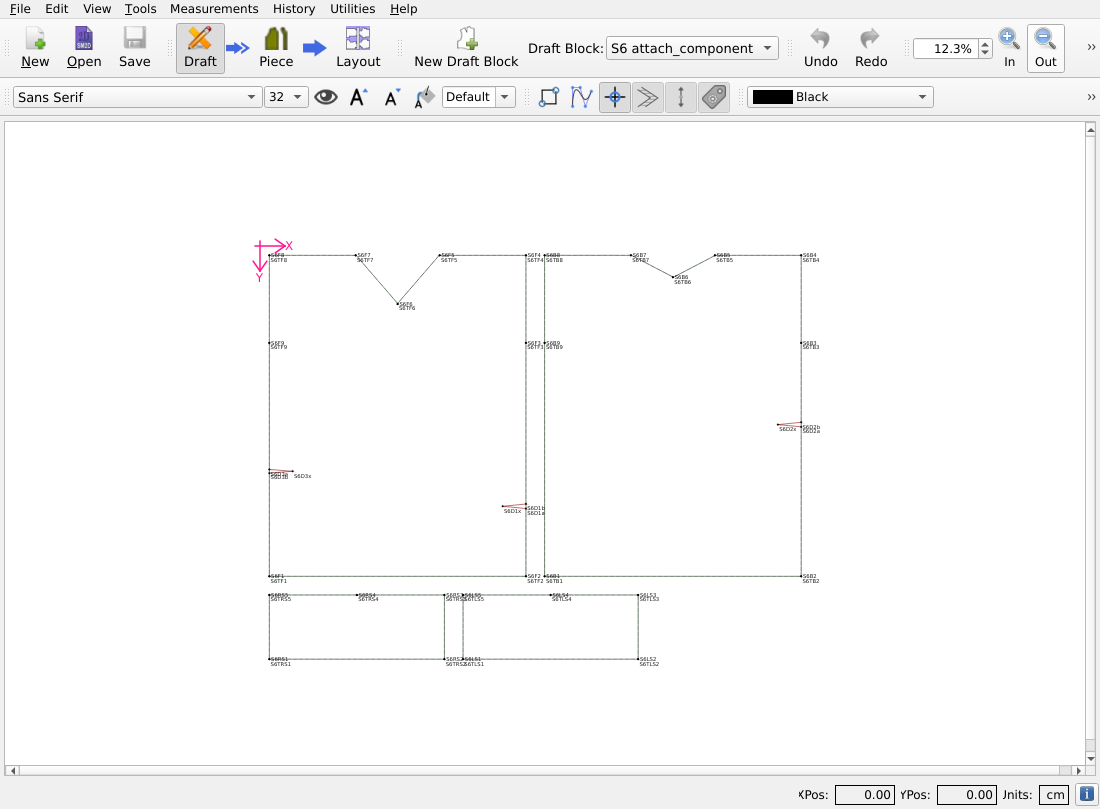}
\caption{A certified trajectory replayed in Seamly2D: base state with the target overlaid as dashed contours (left) and the certified endpoint (right).}
\label{fig:seamly}
\end{figure}

To show that certified programs are usable outside the environment, we translated a certified Tier-C trajectory into the open-source drafting tool Seamly2D \citep{seamly2d}: each intermediate state becomes a draft block, the target contour is overlaid, and the tool's own console exporter produces PDF and DXF (Fig.~\ref{fig:seamly}).
A grader that reads Seamly2D files and applies the public tight criterion passes the certified endpoint, rejects the untouched base and a sleeve shortened by 1 cm, and accepts a 0.3 cm perturbation, which places the criterion's tolerance in physical units.

\end{document}